\documentclass[conference]{IEEEtran}
\pdfoutput=1
\IEEEoverridecommandlockouts
\usepackage{cite}
\usepackage{amsmath,amssymb,amsfonts}
\usepackage{algorithmic}
\usepackage{graphicx}
\usepackage{adjustbox}
\usepackage{lscape}
\usepackage{textcomp}
\usepackage{xcolor}
\usepackage{listings}
\usepackage{tcolorbox} 
\usepackage{tabularx}
\usepackage{array}
\usepackage{booktabs} 
\usepackage{multirow} 
\usepackage{makecell}
\usepackage{hyperref}
\usepackage{float}
\usepackage{amsmath}
\usepackage{placeins}

\definecolor{red}{HTML}{de2d26}
\definecolor{green}{HTML}{31a354}

\begin{document}

\title{Advanced Deep Learning Techniques for Classifying Dental Conditions Using Panoramic X-Ray Images\\
}

\author{
\IEEEauthorblockN{Alireza Golkarieh\textsuperscript{1}, 
Kiana Kiashemshaki\textsuperscript{2}, 
Sajjad Rezvani Boroujeni\textsuperscript{3}}
\IEEEauthorblockA{\textsuperscript{1}PhD Student in Computer Science and Informatics, 
Oakland University, Rochester, Michigan, USA}
\IEEEauthorblockA{\textsuperscript{2}Master of Science in Computer Science, 
Department of Computer Science,\\ Bowling Green State University, Ohio, USA}
\IEEEauthorblockA{\textsuperscript{3}Master of Science in Applied Statistics, 
Department of Applied Statistics and Operations Research,\\ Bowling Green State University, Ohio, USA}
\IEEEauthorblockA{Emails: golkarieh@oakland.edu, kkiana@bgsu.edu, sajjadr@bgsu.edu}
}

\maketitle

\begin{abstract}
This study aimed to develop and evaluate multiple deep learning approaches for automated classification of dental conditions in panoramic radiographs, comparing the performance of custom convolutional neural networks (CNNs), hybrid CNN-machine learning models, and fine-tuned pre-trained architectures for detecting fillings, cavities, implants, and impacted teeth. A dataset of 1,512 panoramic dental X-ray images containing 11,137 annotations across four dental conditions was employed, with class imbalance addressed through random downsampling to create a balanced dataset of 894 samples per condition. Multiple computational approaches were implemented and evaluated using 5-fold cross-validation, including a custom CNN architecture, hybrid models combining CNN feature extraction with traditional machine learning classifiers (Support Vector Machine, Decision Tree, and Random Forest), and three fine-tuned pre-trained architectures (VGG16, Xception, and ResNet50). Performance evaluation was conducted using standard classification metrics including accuracy, precision, recall, and F1-score. The hybrid CNN-Random Forest model achieved the highest performance with 85.4$\pm$2.3\% accuracy, representing an 11 percentage point improvement over the custom CNN baseline (74.29\%). Among pre-trained architectures, VGG16 demonstrated superior performance with 82.3$\pm$2.0\% accuracy, followed by Xception (80.9$\pm$2.3\%) and ResNet50 (79.5$\pm$2.7\%). The CNN+Random Forest model exhibited exceptional performance for fillings detection (F1-score: 0.860$\pm$0.033) and maintained balanced classification across all dental conditions. Systematic misclassification patterns were observed between morphologically similar conditions, particularly cavity-implant and cavity-impacted tooth categories, highlighting the inherent challenges in distinguishing overlapping dental pathologies. Hybrid CNN-based approaches, particularly the combination of CNN feature extraction with Random Forest classification, provide enhanced discriminative capability for automated dental condition detection compared to standalone architectures. The computational efficiency and superior performance of hybrid models demonstrate their potential for clinical deployment as supportive diagnostic tools. However, the observed misclassification patterns between morphologically similar conditions indicate that these AI systems should function as adjunct rather than replacement tools for clinical expertise, requiring further validation through prospective clinical studies.
\end{abstract}


\section{Introduction}

Dental health plays a crucial role in overall human well-being, affecting essential functions such as mastication, speech articulation, facial aesthetics, and psychological confidence, while poor oral health has been linked to systemic conditions including cardiovascular disease, diabetes, and respiratory infections \cite{b1}. According to the World Health Organization, oral diseases affect nearly 3.5 billion people worldwide, with untreated dental caries being the most common health condition globally, affecting approximately 2.3 billion people \cite{b2}. The economic burden of dental diseases is substantial, with treatment costs exceeding \$442 billion annually worldwide, emphasizing the critical need for early detection and preventive interventions \cite{b3}. Traditional diagnostic methods for dental condition assessment, particularly through X-ray radiographic imaging, while widely utilized in clinical practice, are heavily dependent on clinician expertise and experience, leading to potential variability in diagnostic accuracy and increased examination time. With the exponential growth of dental X-ray imaging data and the emergence of artificial intelligence in healthcare, deep learning techniques have demonstrated remarkable potential for automated medical image analysis and pattern recognition in radiographic diagnostics \cite{b4}. The integration of deep learning algorithms in X-ray-based dental diagnostics offers unprecedented opportunities to enhance diagnostic precision, reduce human error, and provide consistent, objective assessments of various dental pathologies including caries, periodontal diseases, and anatomical anomalies. Given the increasing global burden of dental diseases and the growing demand for efficient diagnostic tools, the development of automated classification systems using deep learning methodologies for X-ray image analysis has become not only beneficial but essential for advancing dental healthcare delivery and improving patient care standards worldwide.

The application of artificial intelligence in dental diagnostics has witnessed remarkable advancement over the past decade, with numerous research investigations demonstrating the potential of machine learning and deep learning methodologies for automated classification of various dental conditions through X-ray image analysis. Early pioneering studies established foundational frameworks for dental image processing, with researchers exploring traditional machine learning approaches before transitioning to more sophisticated deep neural network architectures \cite{b4}. Contemporary investigations have focused extensively on the detection and classification of dental caries, representing the most prevalent oral pathology globally. Studies utilizing convolutional neural networks (CNNs) for caries detection have consistently achieved impressive performance metrics, with Lee et al. reporting detection accuracy of 82\% using deep CNN architectures on periapical radiographs \cite{b5}.

Recent research has expanded beyond simple caries detection to encompass comprehensive classification schemes for multiple dental conditions. A comprehensive study by Rahman et al. established a standardized dental dataset categorizing panoramic radiographs into six major classes: healthy teeth, caries, impacted teeth, infections, fractured teeth, and broken-down crowns/roots, achieving classification accuracies ranging from 85\% to 93\% across different deep learning models \cite{b6}. Performance metrics from various studies reveal encouraging results across different dental pathologies and imaging modalities. Research conducted by Takahashi et al. on cone beam computed tomography (CBCT) images demonstrated that deep learning algorithms could accurately detect and classify dental caries with sensitivity rates of 89.7\% and specificity of 92.3\% \cite{b7}. Similarly, investigations utilizing U-Net architectures for early caries detection in bitewing radiographs achieved area under the curve (AUC) values of 0.91, significantly improving clinician diagnostic performance \cite{b8}. The evolution of deep learning methodologies in dental diagnostics has progressed from binary classification tasks to multi-class problems addressing complex diagnostic scenarios. Particularly noteworthy are studies focusing on dental implant classification, where researchers have achieved remarkable accuracy rates. A landmark investigation by Kurt Bayrakdar et al. utilized transfer learning with VGG16 and VGG19 models to classify 11 different dental implant systems from panoramic X-ray images, achieving classification accuracies of 99.04\% and 98.48\% respectively using 8,859 implant images \cite{b9}. Recent work by Kim et al. demonstrated that deep learning models could classify dental implant diameter and length with 95.6\% accuracy using periapical radiographs, while clustering analysis approaches achieved 87.3\% accuracy for the same classification task \cite{b10}. Contemporary research has also addressed periodontal disease detection and classification through artificial intelligence approaches. Studies have successfully implemented various architectural approaches including ResNet variants for feature extraction and ensemble methods combining multiple network architectures to enhance diagnostic reliability across different pathological conditions \cite{b11}. The development of automated systems for filled tooth detection and restorative material classification has shown promising results, with researchers achieving classification accuracies exceeding 90\% for distinguishing between different filling materials and crown types using deep CNN architectures \cite{b12}. Current research trends indicate a shift toward more sophisticated applications including comprehensive oral pathology assessment, with recent systematic reviews highlighting the clinical potential of AI-based diagnostic tools in improving both accuracy and efficiency of dental practice \cite{b13}.

Despite the progress in dental AI research, gaps remain in current methodologies that affect their clinical applicability. Most existing studies have focused on single-pathology detection or binary classification tasks, with limited investigation into simultaneous multi-class classification of different dental conditions within unified frameworks. Furthermore, previous research has primarily emphasized detection algorithms without sufficient attention to entity-based segmentation techniques that provide spatial information about specific dental condition regions. The integration of entity annotation approaches with classification networks for panoramic radiograph analysis requires further exploration, particularly for assessment of multiple dental conditions including fillings, cavities, implants, and impacted teeth within a single diagnostic system.

This study addresses these limitations by developing an integrated approach that combines entity-based segmentation with multi-class classification for dental condition assessment using panoramic X-ray images. Our research methodology employs entity annotation to identify and segment specific dental condition regions within panoramic radiographs, followed by feature extraction through pre-trained CNN architectures and classification using hybrid models that utilize flattened CNN features for discrimination between four dental conditions: fillings, cavities, implants, and impacted teeth. This framework provides both spatial localization through entity segmentation and classification capabilities within a unified system, contributing to improved diagnostic consistency in dental practice. The integration of entity-based segmentation with classification networks offers enhanced accuracy in dental condition assessment, supporting the development of practical AI-based diagnostic tools for clinical applications.

\section{Materials and Methods}

\subsection{Data Collection}

\begin{figure}[!t]
  \centering
  \includegraphics[width=\columnwidth]{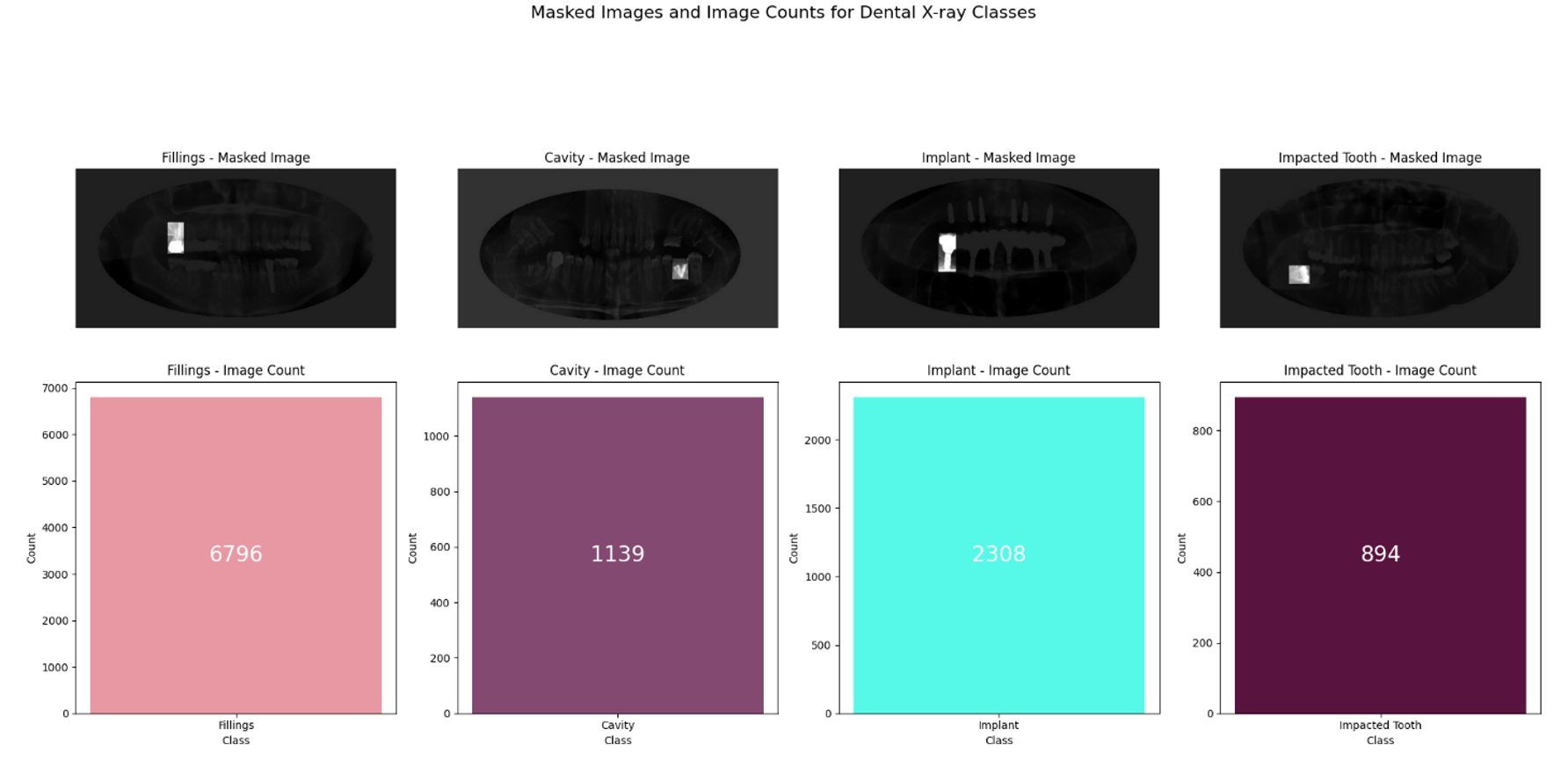}
  \caption{Illustration of Masked Dental X-ray Images and Class Distribution in the Study Dataset}
  \label{fig:dataset}
\end{figure}

The dataset utilized in this study is publicly available and comprises 1,512 panoramic dental X-ray images with a spatial resolution of 512 $\times$ 256 pixels. Each tooth in the images was carefully annotated for the presence of four dental conditions: fillings, cavities, implants, and impacted teeth, resulting in a total of 11,137 annotations. Specifically, the dataset includes 6,797 annotations for fillings, 2,308 for implants, 1,139 for cavities, and 894 for impacted teeth. To ensure the reliability and clinical accuracy of the annotations, all labeled regions were independently reviewed and validated by two experienced dental specialists. Only the annotations confirmed by both experts were considered in the final dataset, thereby enhancing the credibility and robustness of the data used for model development. Representative examples of the annotated radiographs and the class distribution are illustrated in Figure~\ref{fig:dataset}, highlighting the relative frequency of each dental condition.

\FloatBarrier % prevents the next figure from floating above this point

\subsection{Preprocessing}

To ensure the quality and consistency of the input data, a comprehensive preprocessing pipeline was applied to the panoramic dental X-ray images. The goal was to enhance image clarity, reduce noise, and standardize the dataset for reliable feature extraction. The preprocessing steps included brightness adjustment, noise removal, local contrast enhancement, normalization, mask application, and resizing. Each technique was carefully selected to address specific challenges such as uneven illumination, noise artifacts, and inconsistent image dimensions. A summary of the preprocessing methods, including formulas, descriptions, and parameter settings, is provided in Table~\ref{tab:preprocessing}. After mask application, the dataset was balanced to mitigate class imbalance and prevent bias during training. Specifically, random downsampling was performed so that the number of samples in each class was reduced to match the class with the lowest frequency. This ensured equal representation of all four dental conditions in the final dataset. The distribution of samples before and after balancing is presented in Table~\ref{tab:distribution}.

\begin{table*}[htbp]
\centering
\caption{Overview of preprocessing techniques and dataset balancing after mask application.}
\label{tab:preprocessing}
\begin{tabular}{|p{3.5cm}|p{5.2cm}|p{5.5cm}|p{3.2cm}|}
\hline
\textbf{Method} & \textbf{Formula} & \textbf{Description} & \textbf{Parameters} \\
\hline
Brightness Adjustment & $I'(x,y)=\alpha \cdot I(x,y) + \beta$ & Adjusts brightness and contrast to enhance feature visibility. & $\alpha=1.5, \; \beta=15$ \\
\hline
Noise Removal~\cite{b14} & $I'(x,y)=\text{median}(I(x-k,y-k),\dots,I(x+k,y+k))$ & Reduces salt-and-pepper noise while preserving edges using a median blur filter. & Kernel size $k=3$ \\
\hline
Contrast Enhancement~\cite{b15} & $I'=\text{CLAHE}(I)$ & Enhances local contrast using CLAHE, avoiding excessive noise amplification. & clipLimit = 2.0, tileGridSize = $3 \times 3$ \\
\hline
Normalization & $I'=\dfrac{I - \min(I)}{\max(I) - \min(I)}$ & Scales pixel intensities to the range [0,1] for faster neural network convergence. & None \\
\hline
Applying Mask & 
$M(x,y)= 
\begin{cases} 
1, & \text{if $(x,y)$ is within the bounding box}\\ 
0, & \text{otherwise} 
\end{cases}$ 

$I'(x,y)=I(x,y)\cdot M(x,y)$ & Focuses the model on relevant regions by applying bounding box masks. & None \\
\hline
Resizing & $I'(x,y)=I\left(\dfrac{x\cdot width}{224},\dfrac{y\cdot height}{224}\right)$ & Resizes images to a uniform dimension to ensure consistency in input size for the models. & Final size = $224 \times 224$, nearest-neighbor interpolation \\
\hline
Random Downsampling (Balancing) & -- & Randomly reduces the number of samples in each class to match the smallest class size after masking. & Based on class with minimum samples (894) \\
\hline
\end{tabular}
\end{table*}

\begin{table*}[htbp]
\centering
\caption{Class distributions of the four dental conditions before and after balancing.}
\label{tab:distribution} 
\begin{tabular}{|l|c|c|}
\hline
\textbf{Dental Condition} & \textbf{Before Balancing} & \textbf{After Random Downsampling} \\
\hline
Fillings       & 6,797 & 894 \\
Implants       & 2,308 & 894 \\
Cavities       & 1,139 & 894 \\
Impacted Teeth & 894   & 894 \\
\hline
\textbf{Total} & 11,137 & 3,576 \\
\hline
\end{tabular}
\end{table*}

\begin{figure}[!t]
  \centering
  \includegraphics[width=\columnwidth]{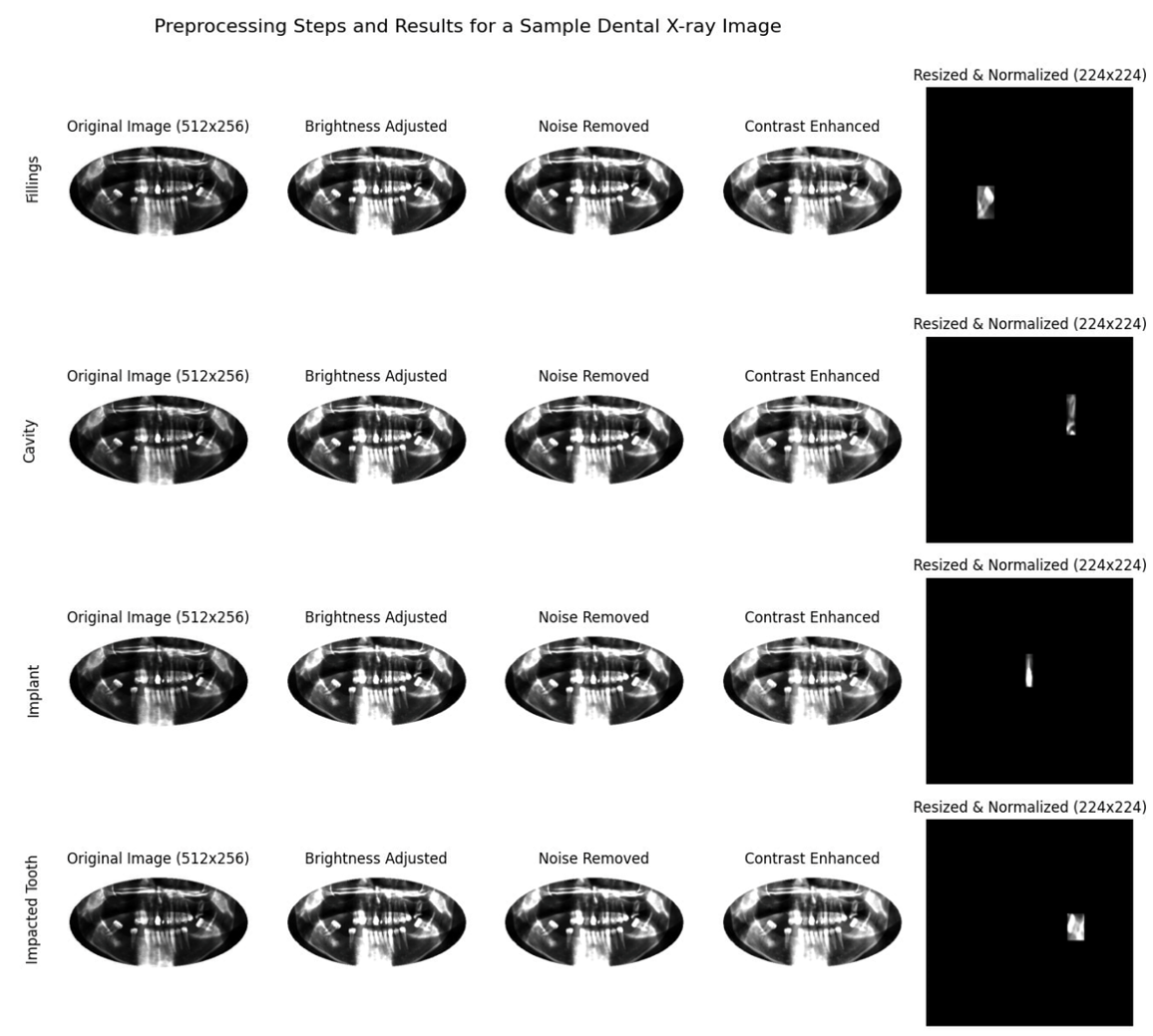}
  \caption{Step-by-step visualization of preprocessing operations, from raw input to the final standardized images.}
  \label{fig:preprocessing}
\end{figure}

\subsection{Model Development}
\subsubsection{Independent CNN Model}
\begin{figure}[!t]
    \centering
    \includegraphics[width=0.85\columnwidth]{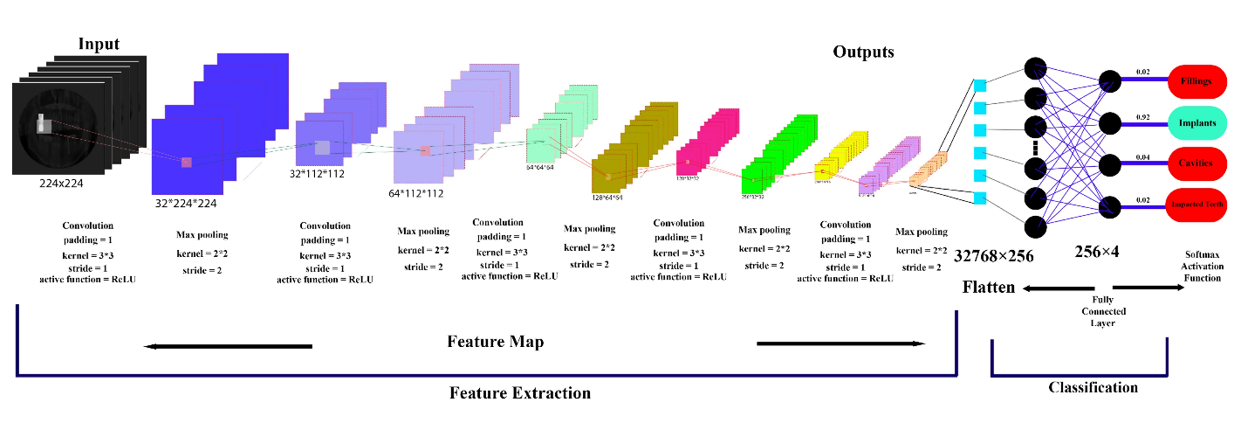}
    \caption{Architecture of a Convolutional Neural Network (CNN) Model}
    \label{fig:cnn_architecture}
\end{figure}

A convolutional neural network (CNN) was developed for classifying dental conditions in panoramic X-ray images. The architecture includes Conv2D layers with 3$\times$3 filters and ReLU activation for feature extraction, MaxPooling layers (2$\times$2) for spatial reduction, and dropout layers to prevent overfitting. The final dense layers perform classification, with the output layer using a softmax activation to generate class probabilities. Model training employed the Adam optimizer and sparse categorical cross-entropy loss, with a batch size of 16, a 10\% validation split, and early stopping to improve efficiency and generalization. The complete architecture and hyperparameters are summarized in Table~\ref{tab:cnn_params}, and Figure~\ref{fig:cnn_architecture} illustrates the CNN workflow for feature extraction and classification.

\subsubsection{Hybrid Models: CNN with SVM, RF, and DT}

In addition to the standalone CNN model, hybrid models were developed by combining the CNN with traditional machine learning classifiers such as Support Vector Machine (SVM), Random Forest (RF), and Decision Tree (DT). These hybrid models leverage the feature extraction capabilities of CNN and the classification strengths of SVM, RF, and DT.

\textbf{CNN Architecture for Feature Extraction} \\
The CNN architecture used for feature extraction is the same as the standalone model, consisting of convolutional layers, pooling layers, and dropout layers. The output of the final dropout layer is flattened to create feature vectors for input into the traditional classifiers.

\textbf{Traditional Classifiers for Classification} \\
The flattened feature vectors from the CNN are used as input to three different classifiers: SVM, RF, and DT. Each classifier is trained to predict whether the input image is cancerous or non-cancerous.

\textbf{Support Vector Machine (SVM)}: SVM is a supervised learning model that finds the hyperplane that best separates the classes \cite{b16}.  

\begin{equation}
f(x) = \text{sign}(w^T x + b)
\end{equation}

where $w$ is the weight vector and $b$ is the bias. The SVM model employs an RBF kernel for classification, with various hyperparameters fine-tuned for optimal performance. Table~\ref{tab:cnn_params} provides a comprehensive overview of these learning parameters.

\textbf{Random Forest (RF)}: RF is an ensemble learning method that uses multiple decision trees to improve classification accuracy \cite{b17}.  

\begin{equation}
f(x) = \frac{1}{T} \sum_{t=1}^{T} h_t(x)
\end{equation}

where $T$ is the number of trees and $h_t$ is the DT classifier. The Random Forest model incorporates 100 estimators to create a robust ensemble classifier. For further details on parameter settings and cross-validation, refer to Table~\ref{tab:cnn_params}.  

\textbf{Decision Tree (DT)}: DT is a supervised learning model that splits the data into branches to make decisions based on the feature values. It works by recursively partitioning the feature space into regions with similar labels.  

\begin{equation}
f(x) = \sum_{i=1}^{n} D(x \in R_i) \cdot c_i
\end{equation}

where $R_i$ are the regions defined by the decision nodes, $D$ is the indicator function, and $c_i$ is the class label assigned to region $R_i$. The Random Forest model incorporates 100 estimators to create a robust ensemble classifier. All learning parameters used are specified in Table~\ref{tab:cnn_params}.

\begin{table*}[htbp]
\centering
\caption{Summary of the CNN Architecture and Learning Parameters for Machine Learning Models}
\label{tab:cnn_params}
\begin{tabular}{|p{2cm}|p{6.5cm}|p{6.5cm}|p{2.5cm}|}
\hline
\textbf{Model} & \textbf{Architecture Details} & \textbf{Learning Parameters} & \textbf{Activation Function} \\
\hline

CNN & 
Input: (224, 224, 1) $\rightarrow$ Conv2D (32, 3$\times$3, ReLU) $\rightarrow$ MaxPooling (2$\times$2) $\rightarrow$ Dropout (0.3) $\rightarrow$ Conv2D (64, 3$\times$3, ReLU) $\rightarrow$ MaxPooling (2$\times$2) $\rightarrow$ Dropout (0.3) $\rightarrow$ Conv2D (128, 3$\times$3, ReLU) $\rightarrow$ MaxPooling (2$\times$2) $\rightarrow$ Dropout (0.3) $\rightarrow$ Conv2D (256, 3$\times$3, ReLU) $\rightarrow$ MaxPooling (2$\times$2) $\rightarrow$ Dropout (0.3) $\rightarrow$ Flatten $\rightarrow$ Dense (256, ReLU) $\rightarrow$ Dropout (0.3) $\rightarrow$ Dense (4, Softmax) & 
Optimizer: Adam \newline
Loss: Sparse Categorical Cross-Entropy \newline
Batch Size: 16 \newline
Epochs: 30 \newline
Dropout: 0.3--0.5 \newline
Batch Normalization: Yes \newline
Cross-Validation: 5-fold &
ReLU (hidden) \newline Softmax (output) \\
\hline

SVM & 
-- & 
Standardize Data: True \newline
Solver: SMO \newline
Cross-Validation: 5 \newline
Kernel: RBF \newline
C: 1.0 \newline
Gamma: `scale` \newline
Probability: True \newline
Cross-Validation: 10 &
-- \\
\hline

DT & 
-- & 
Standardize Data: True \newline
Criterion: `gini` \newline
Max Depth: None \newline
Min Samples Split: 2 \newline
Estimators: 100 \newline
Cross-Validation: 5 &
-- \\
\hline

RF & 
-- & 
Standardize Data: True \newline
Cross-Validation: 5 \newline
Number of Estimators: 100 \newline
Criterion: `gini` \newline
Max Features: `auto` \newline
Cross-Validation: 5 &
-- \\
\hline

\end{tabular}
\end{table*}

\subsubsection{Fine-Tuning with Pretrained Models}

In addition to the independently designed CNN, three well-established pretrained architectures (VGG16, ResNet50, and Xception) were fine-tuned for the classification of four dental conditions (fillings, cavities, implants, and impacted teeth). The transfer learning approach leverages the representational power of models trained on large-scale datasets (e.g., ImageNet) and adapts them to the domain of dental radiography. Each panoramic X-ray image was resized to 224 $\times$ 224 pixels and replicated across three channels to comply with the input requirements of the pretrained networks. The fully connected classification layers of each model were replaced with task-specific dense layers, followed by a softmax output for four-class prediction. Fine-tuning was performed on the deeper convolutional layers while keeping the earlier feature extractors partially frozen to preserve generic image representations.

\textbf{VGG16:} The VGG16 architecture is characterized by a uniform deep design consisting of stacked convolutional layers with small 3$\times$3 kernels, interleaved with max-pooling operations. This design emphasizes capturing fine spatial patterns and progressively reducing dimensionality. The feature extraction at layer $l$ is defined as \cite{b18}:

\begin{equation}
F_l = \sigma(W_l * F_{l+1} + b_l)
\end{equation}

where $F_l$ is the feature map, $W_l$ and $b_l$ denote the kernel weights and bias, and $\sigma$ is the non-linear ReLU activation. The simplicity and depth of VGG16 make it effective for medical imaging tasks, although the model is computationally intensive due to its large parameter set.

\textbf{ResNet50:} The ResNet50 model introduces \textit{residual connections} that address the vanishing gradient problem in very deep networks. Its fundamental unit is the residual block, formulated as \cite{b19}:

\begin{equation}
F_l = \sigma(W_l * F_{l+1} + b_l) + F_{l-1}
\end{equation}

where the shortcut term ($+F_{l-1}$) enables direct information propagation across layers, facilitating efficient training of networks exceeding 50 layers. Additionally, the architecture employs batch normalization and ReLU activation after convolutional operations, which accelerates convergence and enhances generalization. The ability of ResNet50 to learn hierarchical features with improved gradient flow makes it particularly suitable for complex radiographic data.

\textbf{Xception:} The Xception model builds upon the Inception family by adopting \textit{depthwise separable convolutions} in place of standard convolutions. This decomposition reduces computational complexity while preserving high representational power. The operation at layer $l$ is expressed as \cite{b20}:

\begin{equation}
F_l = \sigma(((W_l^d * F_{l-1}) \otimes W_l^p) + b_l)
\end{equation}

where $W_l^d$ represents depthwise filters, $W_l^p$ the pointwise filters, and $\otimes$ indicates channel-wise combination. This factorization enables the model to independently capture spatial correlations through depthwise filters and cross-channel interactions via pointwise filters. Owing to its efficiency and strong performance, Xception is well-suited for large-scale medical image analysis.

\begin{figure*}[!t]
    \centering
    \includegraphics[width=\columnwidth]{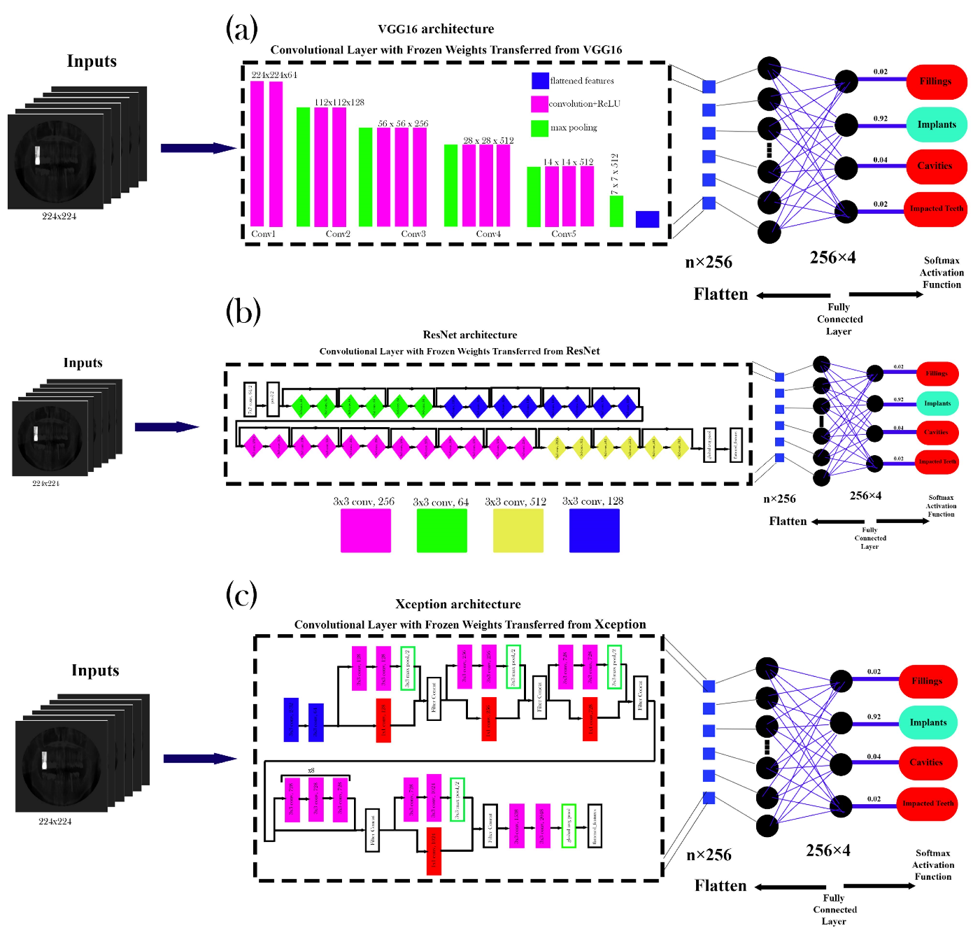}
    \caption{Fine-tuned pretrained architectures for dental condition classification.}
    \label{fig:pretrained_models}
\end{figure*}

Figure~\ref{fig:pretrained_models} illustrates the architectures of the three pretrained models (VGG16 (a), ResNet50 (b), and Xception (c)) used for fine-tuning on panoramic dental X-ray images. Each model receives input images resized to 224$\times$224 pixels and duplicated across three channels to match the pretrained network requirements. The convolutional layers with frozen weights extract hierarchical features, which are then flattened and passed through fully connected layers for classification into four dental conditions. A softmax activation function is applied in the output layer to produce class probabilities.

The learning parameters (optimizer, loss function, batch size, number of epochs, and training strategies) are summarized in Table~\ref{tab:pretrained_params}. By combining domain-specific fine-tuning with the robust architectures of VGG16, ResNet50, and Xception, the models were optimized to achieve accurate and reliable classification of dental conditions from X-ray images.

\begin{table*}[htbp]
\centering
\caption{Architectural Configurations and Training Hyperparameters of Fine-Tuned Models}
\label{tab:pretrained_params}
\begin{tabular}{|p{2.5cm}|p{7cm}|p{7cm}|}
\hline
\textbf{Model} & \textbf{Architecture Details} & \textbf{Learning Parameters} \\
\hline

\textbf{ResNet50} & 
Input: (224 $\times$ 224 $\times$ 3), Pretrained on ImageNet, GlobalAveragePooling2D, Dense: 256, Frozen Layers: all except last block, 5-fold cross-validation &
Optimizer: Adam, Learning Rate = 0.0001, Batch Size = 8, Epochs = 30, Early Stopping (patience = 5); Loss Function: Sparse Categorical Cross-Entropy; Activation: ReLU (hidden), Softmax (output) \\
\hline

\textbf{Xception} & 
Input: (224 $\times$ 224 $\times$ 3), Pretrained on ImageNet, GlobalAveragePooling2D, Dense: 256, Frozen Layers: all except last block, 5-fold cross-validation &
Optimizer: Adam, Learning Rate = 0.0001, Batch Size = 8, Epochs = 30, Early Stopping (patience = 5); Loss Function: Sparse Categorical Cross-Entropy; Activation: ReLU (hidden), Softmax (output) \\
\hline

\textbf{VGG16} & 
Input: (224 $\times$ 224 $\times$ 3), Pretrained on ImageNet, GlobalAveragePooling2D, Dense: 256, Frozen Layers: all except last block, 5-fold cross-validation &
Optimizer: Adam, Learning Rate = 0.0001, Batch Size = 8, Epochs = 30, Early Stopping (patience = 5); Loss Function: Sparse Categorical Cross-Entropy; Activation: ReLU (hidden), Softmax (output) \\
\hline

\end{tabular}
\end{table*}

\begin{figure*}[!t]
    \centering
    \includegraphics[width=\columnwidth]{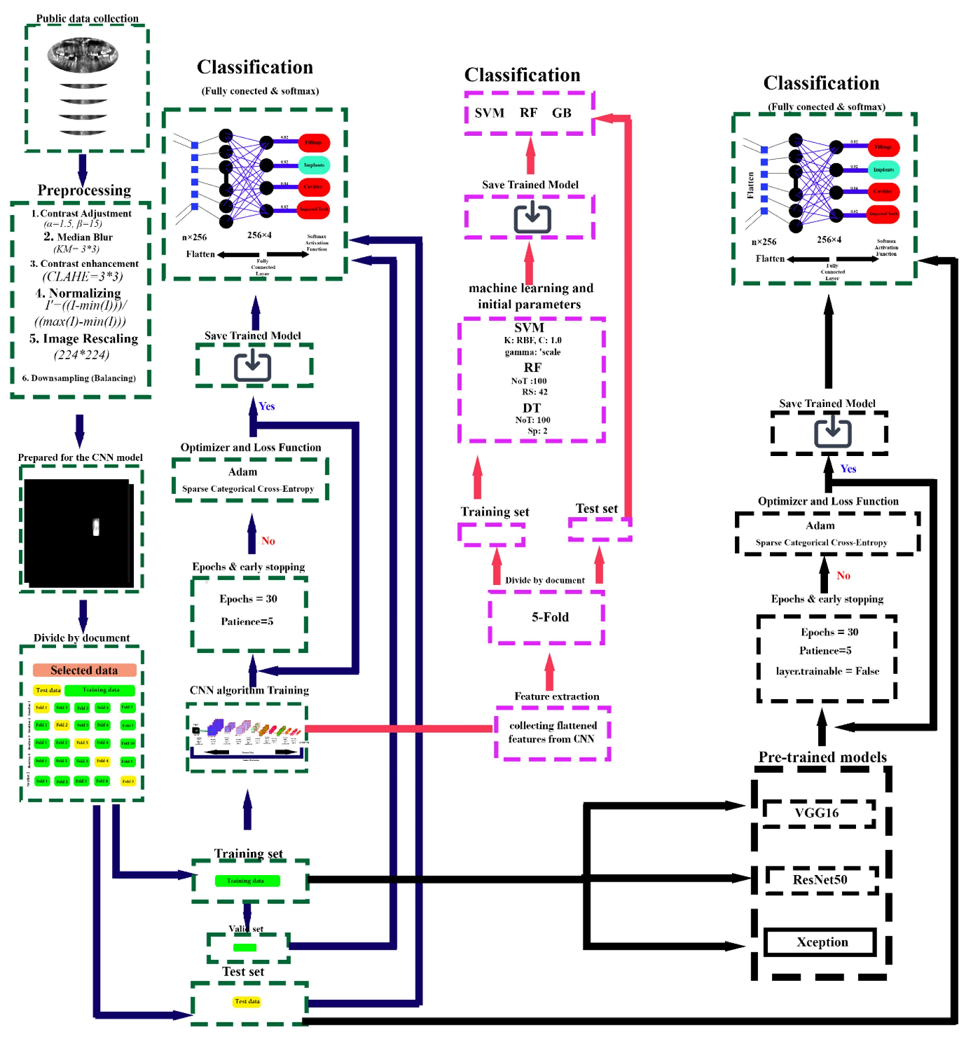}
    \caption{Workflow of the proposed classification framework for dental radiographs.}
    \label{fig:workflow}
\end{figure*}

Figure~\ref{fig:workflow} illustrates the overall methodology, including data preprocessing, CNN-based feature extraction, fine-tuning of pretrained models (VGG16, ResNet50, Xception), and classification using both deep learning and machine learning approaches (SVM, RF, GB). Training, validation, and testing procedures with cross-validation and parameter optimization are also depicted.

\section{Model Evaluation}
\subsection{Performance Metrics}

The evaluation of the proposed classification framework was conducted using widely adopted metrics, namely accuracy, precision, recall, and F1-score, together with the confusion matrix to provide class-wise performance insights.

\begin{itemize}
    \item \textbf{Accuracy} reflects the overall correctness of the model by measuring the proportion of correctly classified samples relative to the total number of samples:
    \begin{equation}
    Accuracy = \frac{TP + TN}{TP + TN + FP + FN}
    \end{equation}

    \item \textbf{Precision} assesses the reliability of positive predictions, indicating the fraction of true positives among all predicted positives:
    \begin{equation}
    Precision = \frac{TP}{TP + FP}
    \end{equation}

    \item \textbf{Recall (Sensitivity)} quantifies the model’s ability to correctly identify actual positive cases:
    \begin{equation}
    Recall = \frac{TP}{TP + FN}
    \end{equation}

    \item \textbf{F1-score} represents the harmonic mean of precision and recall, offering a balanced measure between these two metrics:
    \begin{equation}
    F1\ Score = \frac{Precision \cdot Recall}{Precision + Recall}
    \end{equation}
\end{itemize}

Where $TP$, $TN$, $FP$, and $FN$ denote true positives, true negatives, false positives, and false negatives, respectively.

\subsection{Cross-Validation}

\textbf{k-Fold Cross-Validation:} To obtain a robust and unbiased estimate of the model performance, a 5-fold cross-validation scheme was employed. In this approach, the dataset was partitioned into five equal subsets. During each iteration, one subset was used as the test set, while the remaining four subsets were utilized for training. The process was repeated five times, and the final performance scores were reported as the average across all folds.

\section{Results}

\subsection{Dataset Characteristics and Preprocessing}
A publicly available dataset consisting of 1,512 panoramic dental X-ray images with a resolution of 512 $\times$ 256 pixels was employed in this study. A total of 11,137 annotations were provided for four dental conditions fillings (6,797), implants (2,308), cavities (1,139), and impacted teeth (894). All annotations were independently reviewed and validated by two dental specialists to ensure clinical accuracy. To enhance image quality and maintain consistency, a preprocessing pipeline was implemented, including brightness adjustment, noise reduction, local contrast enhancement, normalization, mask application, and resizing to 224 $\times$ 224 pixels. Class imbalance was addressed through random downsampling, resulting in a balanced dataset of 894 samples per class.

\subsection{Custom CNN Model Performance Evaluation}
\begin{figure*}[!t]
    \centering
    \includegraphics[width=\columnwidth]{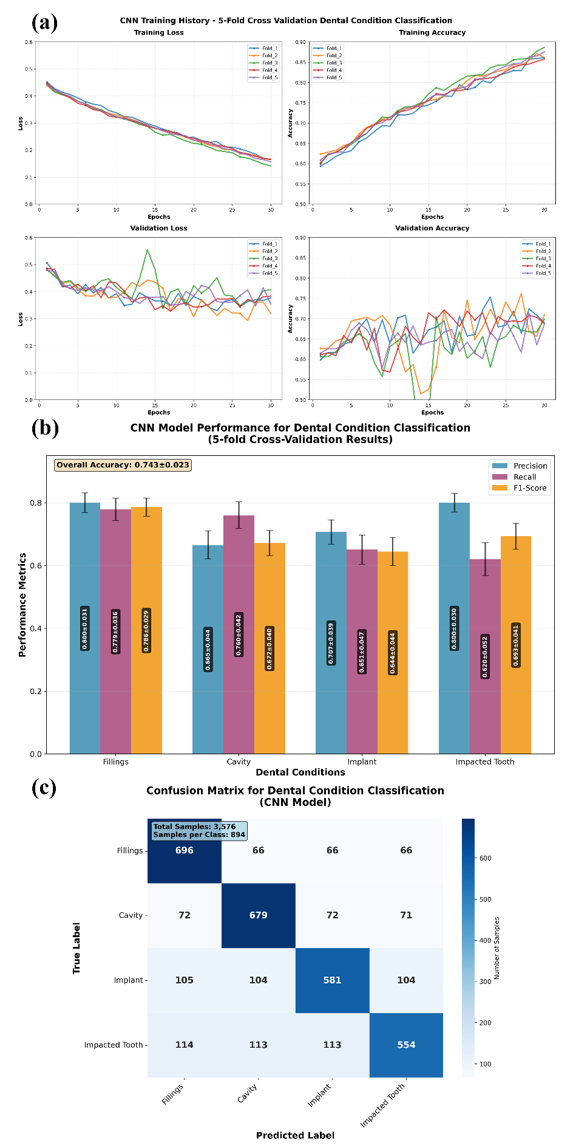}
    \caption{Performance evaluation of custom CNN model for dental condition classification using 5-fold cross-validation.}
    \label{fig:cnn_results}
\end{figure*}

Figure~\ref{fig:cnn_results} presents a comprehensive evaluation of the custom CNN model's performance in classifying four dental conditions (fillings, cavity, implant, and impacted tooth) through 5-fold cross-validation analysis. The evaluation includes training dynamics, classification metrics, and detailed error analysis to assess model reliability and generalization capability.

(a) Training history displaying loss and accuracy curves for both training and validation sets across 30 epochs, with separate lines for each fold. (b) Performance metrics including precision, recall, and F1-score for four dental conditions (fillings, cavity, implant, and impacted tooth) with error bars representing standard deviation across folds. (c) Confusion matrix aggregated from all test samples across the 5-fold cross-validation, showing the distribution of true labels versus predicted labels for the four dental condition classes.

Table~\ref{tab:cnn_results} presents the average performance metrics of the CNN model for dental condition classification across five folds, summarizing the results shown in Figure~\ref{fig:cnn_results}b.

\begin{table*}[htbp]
\centering
\caption{Average Performance Metrics of CNN for Dental Condition Classification Across Five Folds}
\label{tab:cnn_results}
\begin{tabular}{|l|c|c|c|c|}
\hline
\textbf{Dental Condition} & \textbf{Accuracy} & \textbf{Precision} & \textbf{Recall} & \textbf{F1-Score} \\
\hline
Fillings       & 0.7429 & 0.8001 & 0.7787 & 0.7861 \\
Cavity         & 0.7429 & 0.6655 & 0.7602 & 0.6716 \\
Implant        & 0.7429 & 0.7069 & 0.6508 & 0.6445 \\
Impacted Tooth & 0.7429 & 0.8001 & 0.6202 & 0.6930 \\
\hline
\end{tabular}
\end{table*}

The results presented in Figure~\ref{fig:cnn_results} and summarized in Table~\ref{tab:cnn_results} demonstrate that the custom CNN model achieved an overall accuracy of 74.29\% across all dental conditions. The model exhibited varying performance across different conditions, with fillings showing the most balanced performance (precision: 0.80, recall: 0.78, F1-score: 0.79), while impacted teeth demonstrated challenges in recall (0.62) despite high precision (0.80). The confusion matrix analysis of 3,576 total test samples (894 per class) reveals that fillings achieved the highest correct classification rate (696/894), followed by cavity (679/894), implant (581/894), and impacted tooth (554/894). Notable misclassification patterns include 104--105 samples misclassified between cavity and implant categories in both directions, and 113--114 samples misclassified between cavity and impacted tooth, suggesting morphological similarities that challenge the model's discriminative capability. The training curves indicate stable convergence without significant overfitting, demonstrating the model's ability to generalize effectively across different data folds.

\subsection{Hybrid CNN-Based Model Comparison}
\begin{figure*}[!t]
    \centering
    \includegraphics[width=\columnwidth]{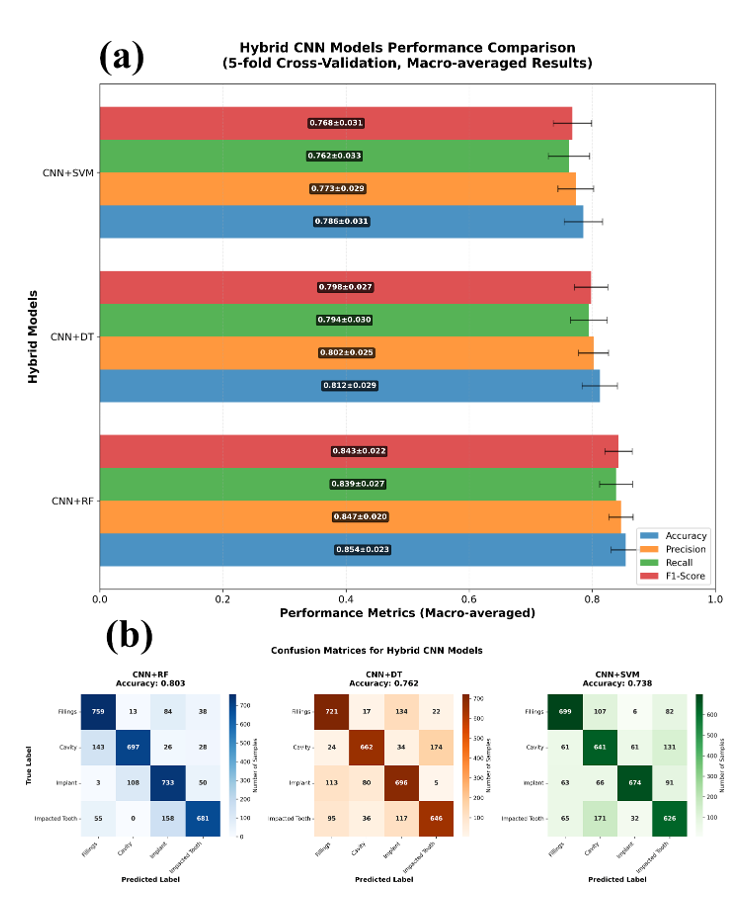}
    \caption{Comparative performance analysis of hybrid CNN-based models for dental condition classification.}
    \label{fig:hybrid_results}
\end{figure*}

Figure~\ref{fig:hybrid_results} presents the performance evaluation of three hybrid models that combine the custom CNN's flattened feature representations with traditional machine learning classifiers (Support Vector Machine, Decision Tree, and Random Forest) for enhanced dental condition detection. The comparison demonstrates how different classification algorithms perform when utilizing CNN-extracted features as input for final prediction decisions.

(a) Macro-averaged performance metrics (accuracy, precision, recall, and F1-score) for three hybrid models across 5-fold cross-validation, with error bars indicating standard deviation. (b) Confusion matrices for each hybrid model showing the classification results across four dental conditions (fillings, cavity, implant, and impacted tooth), with numerical values representing the count of samples in each prediction category.

Table~\ref{tab:hybrid_results} provides the detailed performance metrics of the hybrid CNN-based models for each of the four dental conditions, highlighting the differences across Decision Tree, Random Forest, and SVM classifiers.

\begin{table*}[htbp]
\centering
\caption{Performance Comparison of Hybrid CNN-Based Models for Dental Condition Detection}
\label{tab:hybrid_results}
\begin{tabular}{|l|l|c|c|c|c|}
\hline
\textbf{Model} & \textbf{Dental Condition} & \textbf{F1-Score} & \textbf{Precision} & \textbf{Recall} & \textbf{Accuracy} \\
\hline
\multirow{4}{*}{CNN+DT} 
& Cavity         & 0.793 $\pm$ 0.028 & 0.789 $\pm$ 0.019 & 0.798 $\pm$ 0.026 & 0.812 $\pm$ 0.029 \\
& Fillings       & 0.823 $\pm$ 0.022 & 0.834 $\pm$ 0.032 & 0.812 $\pm$ 0.024 & 0.812 $\pm$ 0.029 \\
& Impacted Tooth & 0.788 $\pm$ 0.025 & 0.784 $\pm$ 0.018 & 0.792 $\pm$ 0.026 & 0.812 $\pm$ 0.029 \\
& Implant        & 0.788 $\pm$ 0.030 & 0.801 $\pm$ 0.024 & 0.776 $\pm$ 0.026 & 0.812 $\pm$ 0.029 \\
\hline
\multirow{4}{*}{CNN+RF} 
& Cavity         & 0.828 $\pm$ 0.021 & 0.823 $\pm$ 0.027 & 0.834 $\pm$ 0.023 & 0.854 $\pm$ 0.023 \\
& Fillings       & 0.860 $\pm$ 0.033 & 0.876 $\pm$ 0.022 & 0.845 $\pm$ 0.039 & 0.854 $\pm$ 0.023 \\
& Impacted Tooth & 0.854 $\pm$ 0.037 & 0.848 $\pm$ 0.029 & 0.859 $\pm$ 0.020 & 0.854 $\pm$ 0.023 \\
& Implant        & 0.828 $\pm$ 0.030 & 0.839 $\pm$ 0.016 & 0.817 $\pm$ 0.037 & 0.854 $\pm$ 0.023 \\
\hline
\multirow{4}{*}{CNN+SVM} 
& Cavity         & 0.749 $\pm$ 0.019 & 0.745 $\pm$ 0.025 & 0.754 $\pm$ 0.032 & 0.786 $\pm$ 0.031 \\
& Fillings       & 0.787 $\pm$ 0.022 & 0.798 $\pm$ 0.024 & 0.776 $\pm$ 0.036 & 0.786 $\pm$ 0.031 \\
& Impacted Tooth & 0.759 $\pm$ 0.034 & 0.748 $\pm$ 0.034 & 0.771 $\pm$ 0.039 & 0.786 $\pm$ 0.031 \\
& Implant        & 0.755 $\pm$ 0.019 & 0.762 $\pm$ 0.027 & 0.748 $\pm$ 0.023 & 0.786 $\pm$ 0.031 \\
\hline
\end{tabular}
\end{table*}

The comparative analysis presented in Figure~\ref{fig:hybrid_results} and detailed in Table~\ref{tab:hybrid_results} reveals significant performance improvements achieved through hybrid CNN-based approaches compared to the standalone custom CNN model. The CNN+RF hybrid model demonstrated superior performance with the highest accuracy of 85.4$\pm$2.3\%, followed by CNN+DT (81.2$\pm$2.9\%) and CNN+SVM (78.6$\pm$3.1\%). Notably, the CNN+RF model achieved exceptional performance for fillings detection (F1-score: 0.860$\pm$0.033, precision: 0.876$\pm$0.022), representing a substantial improvement over the baseline CNN model's 78.61\% F1-score for the same condition. The confusion matrices indicate that CNN+RF achieved the most balanced classification across all dental conditions, with fillings showing the highest correct classification rate (759/894), while CNN+SVM exhibited the most conservative predictions with higher precision but lower recall rates. These results demonstrate that Random Forest effectively leverages CNN-extracted features to enhance discriminative capability, particularly in distinguishing between morphologically similar dental pathologies that challenged the standalone CNN model.

\subsection{Pre-trained Model Evaluation}
\begin{figure*}[!t]
    \centering
    \includegraphics[width=\columnwidth]{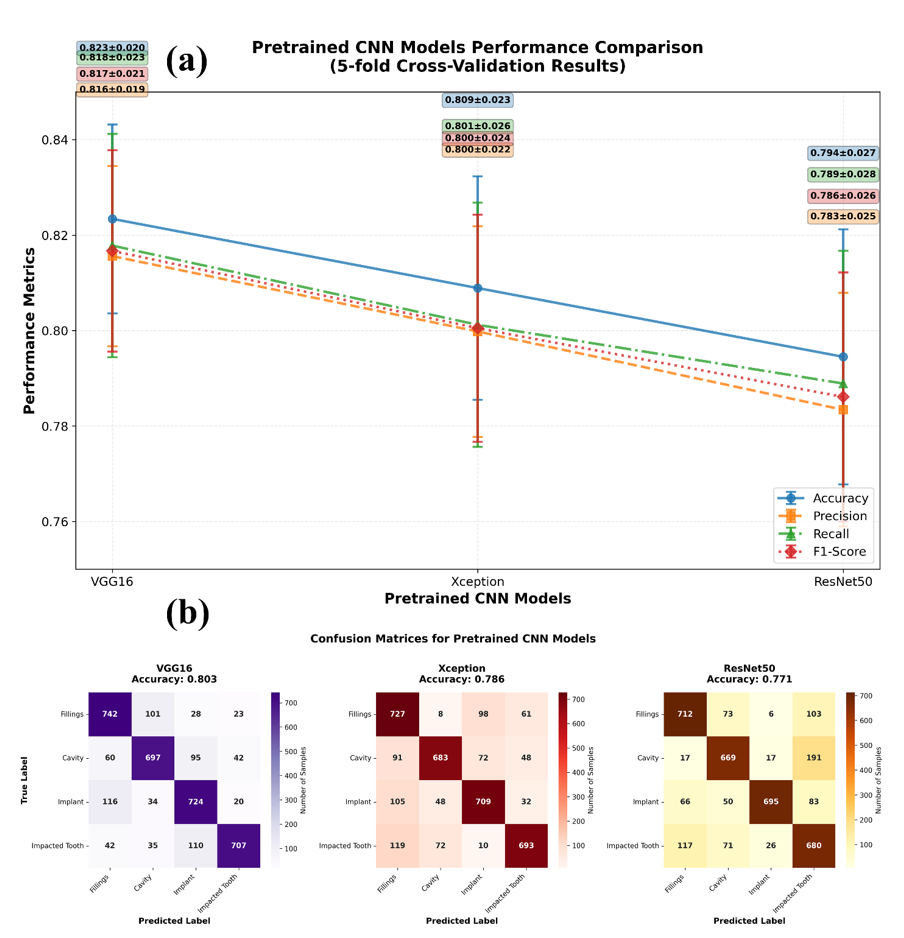}
    \caption{Comparative performance analysis of fine-tuned pre-trained CNN models for dental condition classification.}
    \label{fig:pretrained_results}
\end{figure*}

Figure~\ref{fig:pretrained_results} presents the performance evaluation of three fine-tuned pre-trained deep learning architectures (VGG16, Xception, and ResNet50) adapted for dental condition detection. The comparison demonstrates how established CNN architectures perform when fine-tuned on dental radiographic data, leveraging transfer learning to enhance classification accuracy across four dental conditions.

(a) Performance metrics comparison across three pre-trained models showing accuracy, precision, recall, and F1-score with error bars representing standard deviation from 5-fold cross-validation. (b) Confusion matrices for each pre-trained model displaying classification results across four dental conditions (fillings, cavity, implant, and impacted tooth), with numerical values indicating the distribution of predicted versus true labels. 

Table~\ref{tab:pretrained_results_table} summarizes the performance comparison of three pre-trained deep learning models (VGG16, Xception, and ResNet50) for the detection of four dental conditions, providing detailed metrics across F1-score, precision, recall, and accuracy.

\begin{table*}[htbp]
\centering
\caption{Performance Comparison of Pre-trained Deep Learning Models for Dental Condition Detection}
\label{tab:pretrained_results_table}
\begin{tabular}{|l|l|c|c|c|c|}
\hline
\textbf{Model} & \textbf{Dental Condition} & \textbf{F1-Score} & \textbf{Precision} & \textbf{Recall} & \textbf{Accuracy} \\
\hline
\multirow{4}{*}{\textbf{VGG16}} 
& Cavity         & 0.808 $\pm$ 0.026 & 0.801 $\pm$ 0.026 & 0.815 $\pm$ 0.034 & 0.823 $\pm$ 0.020 \\
& Fillings       & 0.831 $\pm$ 0.023 & 0.834 $\pm$ 0.028 & 0.828 $\pm$ 0.027 & 0.823 $\pm$ 0.020 \\
& Impacted Tooth & 0.822 $\pm$ 0.031 & 0.815 $\pm$ 0.023 & 0.830 $\pm$ 0.027 & 0.823 $\pm$ 0.020 \\
& Implant        & 0.805 $\pm$ 0.027 & 0.812 $\pm$ 0.032 & 0.798 $\pm$ 0.033 & 0.823 $\pm$ 0.020 \\
\hline
\multirow{4}{*}{\textbf{Xception}} 
& Cavity         & 0.791 $\pm$ 0.023 & 0.785 $\pm$ 0.028 & 0.798 $\pm$ 0.025 & 0.809 $\pm$ 0.023 \\
& Fillings       & 0.815 $\pm$ 0.026 & 0.818 $\pm$ 0.024 & 0.812 $\pm$ 0.023 & 0.809 $\pm$ 0.023 \\
& Impacted Tooth & 0.804 $\pm$ 0.029 & 0.798 $\pm$ 0.030 & 0.811 $\pm$ 0.034 & 0.809 $\pm$ 0.023 \\
& Implant        & 0.791 $\pm$ 0.030 & 0.798 $\pm$ 0.025 & 0.784 $\pm$ 0.031 & 0.809 $\pm$ 0.023 \\
\hline
\multirow{4}{*}{\textbf{ResNet50}} 
& Cavity         & 0.774 $\pm$ 0.029 & 0.768 $\pm$ 0.021 & 0.781 $\pm$ 0.027 & 0.795 $\pm$ 0.027 \\
& Fillings       & 0.798 $\pm$ 0.025 & 0.801 $\pm$ 0.028 & 0.795 $\pm$ 0.027 & 0.795 $\pm$ 0.027 \\
& Impacted Tooth & 0.797 $\pm$ 0.025 & 0.783 $\pm$ 0.025 & 0.812 $\pm$ 0.029 & 0.795 $\pm$ 0.027 \\
& Implant        & 0.774 $\pm$ 0.026 & 0.781 $\pm$ 0.019 & 0.767 $\pm$ 0.029 & 0.795 $\pm$ 0.027 \\
\hline
\end{tabular}
\end{table*}

The comparative analysis presented in Figure~\ref{fig:pretrained_results} and detailed in Table~\ref{tab:pretrained_results_table} demonstrates that VGG16 achieved the highest overall performance among the pre-trained models with an accuracy of 82.3$\pm$2.0\%, followed by Xception (80.9$\pm$2.3\%) and ResNet50 (79.5$\pm$2.7\%). VGG16 exhibited the most consistent performance across all dental conditions, with particularly strong results for fillings detection (F1-score: 0.831$\pm$0.023, precision: 0.834$\pm$0.028). The confusion matrix analysis reveals that VGG16 achieved the highest correct classification rates for fillings (742/894), cavity (697/894), and implant (724/894) conditions, while demonstrating balanced performance across all categories. Xception showed competitive performance with relatively lower variance in metrics, achieving strong results for fillings (727 correct classifications) and impacted tooth (693 correct classifications). ResNet50, while showing the lowest overall accuracy, maintained reasonable performance consistency with notable strength in impacted tooth detection (680 correct classifications). These results indicate that VGG16's architectural characteristics, particularly its deep convolutional structure, are well-suited for extracting discriminative features from dental radiographic images, outperforming both the custom CNN model and other pre-trained architectures in this specific medical imaging domain.

\section{Discussion}

This study evaluated multiple deep learning approaches on a comprehensive dataset of 1,512 panoramic dental X-ray images containing 11,137 annotations across four dental conditions (fillings, implants, cavities, and impacted teeth). The results demonstrate that hybrid CNN-based models significantly outperformed standalone approaches, with the CNN+Random Forest combination achieving the highest accuracy of 85.4$\pm$2.3\%, representing a substantial improvement over the custom CNN baseline (74.29\%). Among pre-trained architectures, VGG16 emerged as the most effective with 82.3$\pm$2.0\% accuracy, while the hybrid models consistently showed superior performance in distinguishing between morphologically similar dental pathologies. These findings indicate that combining CNN feature extraction with traditional machine learning classifiers, particularly Random Forest, provides enhanced discriminative capability for automated dental condition detection in panoramic radiographs.

Since studies in this domain have utilized diverse datasets with varying sizes, imaging modalities, and annotation protocols, direct quantitative comparison of performance metrics is not feasible. However, examining recent literature provides valuable context for positioning our findings within the broader landscape of automated dental radiograph analysis. Table~\ref{tab:literature_comparison} presents methodological approaches and reported outcomes from contemporary studies to illustrate the current state of research in dental image classification and detection.

\begin{table*}[htbp]
\centering
\caption{Overview of Recent Methodologies and Reported Performance in Dental Radiograph Analysis}
\label{tab:literature_comparison}
\begin{tabular}{|p{3.2cm}|p{3.3cm}|p{3.6cm}|p{4.2cm}|p{3.0cm}|}
\hline
\textbf{Research Study} & \textbf{Dataset Characteristics} & \textbf{Clinical Focus} & \textbf{Computational Approach} & \textbf{Reported Performance} \\
\hline
Jae-Hong Lee (2023)~\cite{b21} & 11,980 radiographic images & Dental implant system identification & Deep Convolutional Neural Network with professional validation & 95.4\% classification accuracy \\
\hline
Vasdev et al. (2023)~\cite{b22} & 16,000 dental images & Multi-class dental disease detection & Pipeline architecture: AlexNet, ResNet-18, ResNet-34 & AlexNet: 85.2\% accuracy \\
\hline
Muhammad Adnan Hasnain (2023)~\cite{b23} & Not specified & Radiographic dental pathology classification & Transfer learning: ResNet-101, Xception, DenseNet-201, EfficientNet-B0 & EfficientNet-B0: 98.91\% accuracy \\
\hline
Chisako Muramatsu (2023)~\cite{b24} & 100 panoramic radiographs & Tooth detection and classification for dental charting & Object detection network with 4-fold cross-validation & 93.2\% classification performance \\
\hline
Kailai Zhang (2023)~\cite{b25} & 1,000 radiographic images & Individual tooth detection and classification & Hierarchical label tree with cascade network architecture & 95.8\% detection accuracy \\
\hline
W. Park (2023)~\cite{b26} & 150,733 dental images & Implant system classification & Modified ResNet-50 with architectural adaptations & 82\% classification accuracy \\
\hline
L. Toledo Reyes (2023)~\cite{b27} & 639 clinical images & Caries progression prediction & Ensemble methods: Decision Trees, Random Forest, XGBoost & AUC scores exceeding 0.70 \\
\hline
F. Schwendicke (2022)~\cite{b28} & 3,293,252 radiographic samples & AI-assisted caries detection cost-effectiveness & Machine learning-based detection algorithms & 80\% diagnostic accuracy \\
\hline
\textbf{Present Investigation (2025)} & 3,576 balanced samples (1,512 original images) & Multi-condition classification: Fillings, Cavities, Implants, Impacted Teeth & Custom CNN, Hybrid CNN-ML, Pre-trained architectures (VGG16, Xception, ResNet50) & Custom CNN: 74.29\%, CNN+RF: 85.40\%, VGG16: 82.23\% \\
\hline
\end{tabular}
\end{table*}

The custom CNN model achieved an overall accuracy of 74.29\% across four dental conditions, as demonstrated in Figure~\ref{fig:cnn_results} and summarized in Table~\ref{tab:cnn_results}, for automated dental pathology detection from panoramic radiographs. The model exhibited differential performance across conditions, with fillings showing the most balanced metrics (precision: 0.80, recall: 0.78, F1-score: 0.79), while impacted teeth presented classification challenges despite high precision (0.80) but lower recall (0.62). This performance aligns with recent findings in dental radiographic analysis, where CNN-based approaches have shown varying effectiveness depending on the morphological complexity of dental pathologies~\cite{b29}. The confusion matrix in Figure~\ref{fig:cnn_results}(c) revealed systematic misclassifications between cavity-implant and cavity-impacted tooth categories, suggesting that these conditions share similar radiographic features that challenge discriminative capability. This observation is consistent with the inherent difficulties in distinguishing overlapping dental pathologies in panoramic images, where anatomical superimposition and varying image quality can obscure diagnostic features~\cite{b30}.

The hybrid CNN-based models demonstrated performance improvements over the standalone CNN approach, as illustrated in Figure~\ref{fig:hybrid_results} and detailed in Table~\ref{tab:hybrid_results}, with the CNN+RF combination achieving the highest accuracy of 85.4$\pm$2.3\%. This represents an enhancement of approximately 11 percentage points compared to the custom CNN baseline, highlighting the effectiveness of combining deep feature extraction with traditional machine learning classifiers. The macro-averaged performance metrics shown in Figure~\ref{fig:hybrid_results}(a) demonstrate the superiority of the Random Forest hybrid approach across all evaluation metrics. The superior performance of Random Forest in this hybrid configuration can be attributed to its ability to handle high-dimensional feature spaces and reduce overfitting through ensemble averaging~\cite{b31}. The CNN+RF model particularly excelled in fillings detection (F1-score: 0.860$\pm$0.033), as shown in Table~\ref{tab:hybrid_results}, suggesting that Random Forest effectively leveraged the CNN-extracted features to distinguish filling materials from natural tooth structures and other pathological conditions. The confusion matrices presented in Figure~\ref{fig:hybrid_results}(b) reveal that CNN+RF achieved balanced classification across all dental conditions, with fillings showing the highest correct classification rate (759/894). This finding aligns with research demonstrating that hybrid approaches combining CNNs with ensemble methods can achieve improved performance in medical imaging tasks by capitalizing on both deep feature representation and robust classification strategies~\cite{b32}. The balanced performance across all dental conditions observed in the CNN+RF model indicates its potential for clinical deployment, where consistent accuracy across different pathological conditions is crucial for reliable diagnostic support.

The evaluation of fine-tuned pre-trained architectures revealed that VGG16 achieved the best performance (82.3$\pm$2.0\% accuracy), outperforming both Xception (80.9$\pm$2.3\%) and ResNet50 (79.5$\pm$2.7\%). This superiority of VGG16 in dental radiographic classification can be attributed to its deep convolutional architecture with small receptive fields, which proves effective for capturing fine-grained textural features characteristic of dental pathologies in panoramic images~\cite{b6}. The consistent performance of VGG16 across all dental conditions, particularly for fillings detection (F1-score: 0.831$\pm$0.023), demonstrates the effectiveness of transfer learning from natural images to dental radiographs, despite the domain shift between general computer vision datasets and medical imaging~\cite{b5}. However, these results are somewhat lower than reported in some recent dental AI studies, where ResNet architectures achieved $>$98\% accuracy~\cite{b33}, potentially due to differences in dataset characteristics, preprocessing methodologies, and task complexity. The competitive performance of all three pre-trained models suggests that established CNN architectures can be successfully adapted for dental diagnostic tasks, though architectural choice remains important for optimizing performance in specific dental imaging contexts.

The computational models were implemented using Python programming language, specifically leveraging libraries such as TensorFlow and Keras for neural network construction and training. The simulations were conducted on a system equipped with an NVIDIA RTX 3050 Ti laptop GPU with 4GB of VRAM to accelerate the training process. The system specifications include an Intel Core i7 processor, 32 GB of RAM, and a Windows 11 operating system. The integrated development environment (IDE) used was PyCharm. Data preprocessing and analysis were performed using additional libraries like NumPy, pandas, and SciPy. The training times of the models reflect their computational efficiency. Among pre-trained models, Xception trained in 35.41 minutes, ResNet50 in 37.92 minutes, and VGG16 in 38.36 minutes, with VGG16 being the most time-intensive due to its higher parameter count. The custom CNN model demonstrated efficiency with a training time of 32.63 minutes. Combined CNN-machine learning classifiers significantly reduced training times, with CNN-SVM, CNN-RF, and CNN-DT completing in 14.31, 12.22, and 10.47 minutes, respectively, by leveraging pre-extracted CNN features. These findings highlight the balance between accuracy and computational cost in optimizing dental condition detection workflows.

Despite the methodological contributions, this study presents several limitations that warrant consideration. The dataset size of 1,512 panoramic images with 894 balanced samples per class may limit the generalizability of the models to broader clinical populations and diverse imaging conditions. The 224$\times$224 pixel resolution after preprocessing may result in loss of fine anatomical details that could be diagnostically relevant for distinguishing morphologically similar conditions, as evidenced by the misclassification patterns between cavity-implant and cavity-impacted tooth categories observed in the confusion matrices. Additionally, the study focused solely on panoramic radiographs, which may not capture the full spectrum of dental pathologies visible in other imaging modalities such as bitewing or periapical X-rays. Future research should incorporate larger, multi-center datasets with diverse imaging protocols to enhance model robustness and clinical applicability. Advanced preprocessing techniques preserving higher resolution details, multi-modal imaging integration, and the development of explainable AI frameworks to provide clinicians with interpretable diagnostic reasoning could improve clinical adoption. Furthermore, prospective clinical validation studies comparing AI-assisted diagnosis with traditional expert evaluation in real-world settings are essential to establish the true clinical utility and cost-effectiveness of these automated dental classification systems.

The clinical applications of AI-based dental radiographic analysis systems present promising opportunities for enhancing diagnostic workflows in modern dental practice. These automated systems can serve as valuable adjunct tools to support clinicians in detecting dental pathologies including fillings, cavities, implants, and impacted teeth in panoramic radiographs, particularly in busy clinical environments where time constraints may limit thorough examination of all radiographic details. The integration of such AI-driven diagnostic aids into existing picture archiving and communication systems (PACS) could facilitate standardized screening protocols and reduce inter-observer variability in radiographic interpretation. However, the inherent limitations in distinguishing morphologically similar dental conditions necessitate that these AI systems function as supportive rather than replacement tools for clinical expertise. The computational efficiency of hybrid machine learning approaches makes them feasible for deployment in diverse clinical settings, including general practice offices and dental schools where they could serve educational purposes for training dental students and residents. While these AI-assisted diagnostic tools show potential for improving diagnostic consistency and reducing oversight of pathological conditions, their clinical implementation requires careful validation through prospective studies comparing AI-assisted interpretations with expert radiologist assessments to establish appropriate clinical workflows and determine optimal human-AI collaborative frameworks for reliable dental diagnosis.

\section{Conclusion}

Based on the comprehensive evaluation of deep learning approaches for automated dental condition classification in panoramic radiographs, this study demonstrates that hybrid CNN-based models significantly outperform standalone architectures for multi-class dental pathology detection. The CNN+Random Forest combination achieved the highest accuracy of 85.4$\pm$2.3\%, representing an 11 percentage point improvement over the custom CNN baseline (74.29\%), while VGG16 emerged as the most effective pre-trained architecture with 82.3$\pm$2.0\% accuracy. The systematic misclassification patterns observed between morphologically similar conditions, particularly cavity-implant and cavity-impacted tooth categories, highlight the inherent challenges in distinguishing overlapping dental pathologies from panoramic images. These findings indicate that combining deep feature extraction with traditional machine learning classifiers, particularly ensemble methods like Random Forest, provides enhanced discriminative capability for automated dental diagnosis.

The computational efficiency of hybrid approaches, with training times ranging from 10.47 to 14.31 minutes compared to 32--38 minutes for standalone CNN models, demonstrates their practical feasibility for clinical deployment. However, several limitations warrant consideration, including the relatively small dataset size of 1,512 images, potential loss of fine anatomical details due to 224$\times$224 pixel preprocessing, and focus on panoramic radiography alone. Future research should address these limitations through larger multi-center datasets, preservation of higher resolution details, integration of multiple imaging modalities, and development of explainable AI frameworks to enhance clinical interpretability. While these AI-assisted diagnostic tools show promise for improving diagnostic consistency and reducing oversight of pathological conditions in clinical practice, their implementation requires careful validation through prospective studies to establish appropriate human-AI collaborative frameworks that position these systems as supportive rather than replacement tools for clinical expertise.

\section*{Declarations}

\textbf{Competing interests:}  
The authors declare that they have no known competing financial interests or personal relationships that could be perceived as influencing the work reported in this paper.

\textbf{Funding:}  
This study did not receive any specific grant from funding agencies in the public, commercial, or not-for-profit sectors.

\textbf{Ethics approval:}  
This study was conducted in accordance with the principles outlined in the Declaration of Helsinki. Ethical approval was obtained from the Ethics Committee of Tabriz University of Medical Sciences, Tabriz, Iran. Written informed consent was waived as the study did not involve direct patient participation or personal data.

\textbf{Consent to Publish:}  
Not applicable.

\textbf{Authors' contributions:}  
[Please specify individual author contributions here if required by the target journal.]

\textbf{Acknowledgements:}  
We extend our gratitude to the Roboflow Universe platform (\url{https://universe.roboflow.com/yolo-cthel/disease-xaijn}) for providing publicly available datasets that significantly supported our research.

\textbf{Data Availability:}  
The data is publicly accessible at the following link: \url{https://universe.roboflow.com/yolo-cthel/disease-xaijn}.

\end{document}